\documentclass[11pt]{article}

\usepackage{syntaxfest2021}

\usepackage{times}
\usepackage{latexsym}

\usepackage{etoolbox}
\usepackage{hyperref}
\usepackage{graphicx}
\usepackage{color}
\usepackage{lscape} 
\usepackage{tabularx}

\usepackage{rotating}

\newtoggle{anonymous}
\togglefalse{anonymous}

\iftoggle{anonymous}{}
{
\syntaxfestfinalcopy 
\setlength\titlebox{6cm}
}

\title{Dependency distance minimization predicts compression}

\iftoggle{anonymous}{}
{
\author{Ramon Ferrer-i-Cancho \\
  Complexity and Quantitative Linguistics Lab, \\ 
  LARCA Research Group \\
  Departament de Ci\`encies de la Computaci\'o \\
  Universitat Polit\`ecnica de Catalunya \\
  Campus Nord, Edifici Omega\\
  Jordi Girona Salgado 1-3 \\
  08034 Barcelona, Catalonia, Spain \\
  {\tt rferrericancho@cs.upc.edu} \\\And
  Carlos G\'omez-Rodr\'iguez \\
  Universidade da Coru\~na, CITIC\\
  FASTPARSE Lab, LyS Research Group\\
  Departamento de Ciencias \\   
  de la Computaci\'on y Tecnolog\'ias \\ 
  de la Informaci\'on \\
  Facultade de Inform\'atica, Elvi\~{n}a, \\
  15071, A Coru\~na, Spain \\
  {\tt cgomezr@udc.es} \\} 
}

\date{}

\begin{document}

\maketitle

\begin{abstract}
Dependency distance minimization (DDm) is a well-established principle of word order. It has been predicted theoretically that DDm implies compression, namely the minimization of word lengths. This is a second order prediction because it links a principle with another principle, rather than a principle and a manifestation as in a first order prediction. Here we test that second order prediction with a parallel collection of treebanks controlling for annotation style with Universal Dependencies and Surface-Syntactic Universal Dependencies. To test it, we use a recently introduced score that has many mathematical and statistical advantages with respect to the widely used sum of dependency distances. We find that the prediction is confirmed by the new score when word lengths are measured in phonemes, independently of the annotation style, but not when word lengths are measured in syllables. In contrast, one of the most widely used scores, i.e. the sum of dependency distances, fails to confirm that prediction, showing the weakness of raw dependency distances for research on word order. Finally, our findings expand the theory of natural communication by linking two distinct levels of organization, namely syntax (word order) and word internal structure.
\end{abstract}

\section{Introduction}

According to the dependency distance minimization (DDm) principle, the distance between heads and their dependent words in a sentence has to be reduced \cite{Ferrer2004b,Gildea2007a}. The principle has been supported widely by studies whose coverage of languages and families is increasing over time (e.g., \newcite{Liu2008a}, \newcite{Futrell2015a}, \newcite{Futrell2020a}, \newcite{Ferrer2020b}). 
For simplicity, such distance is usually measured in words (see \newcite{Ferrer2014e} for an exception) but it could be measured with more precision in syllables or 
phonemes \cite{Ferrer2014e}. In  that  way,  the distance of  a  dependency  would  be  a function of the length of the words defining the dependency and that of the words in-between. 
For this reason, it was predicted that word lengths  should  be  minimized  to  minimize  dependency  distances \cite{Ferrer2013f,Ferrer2017c}, namely the DDm principle predicts compression, i.e., another principle whereby $L$, the mean word length, has to be minimized. However, to our knowledge, such a prediction has never been tested in spite of its great theoretical importance. First, it is crucial for the construction of a theory of language and other natural communication systems \cite{Semple2021a}. A critical component of a theory are the predictions that it can make. For instance, DDm predicts the scarcity of crossing dependencies \cite{Gomez2016a} and compression of word lengths predicts Zipf's law of abbreviation \cite{Ferrer2019c}. 
These are examples of first order predictions, namely manifestations that are predicted by a certain principle. The focus of this article are second order predictions, namely, principles that are predicted by other principles, such as (a) the prediction of DDm from Dm, a general principle of distance minimization \cite{Ferrer2020b,Ferrer2003i} in the spirit of Behaghel's pioneering views \cite{Behaghel1932a}, or (b) a principle of surprisal or entropy minimization ($Hm$) from compression \cite{Ferrer2015b}. Table \ref{theory_table} summarizes these and other first order and second order predictions from previous research stemming from a general principle of energy minimization (Em) in the spirit of Zipf's least effort principle \cite{Zipf1949a} but extended to other biological systems \cite{Semple2021a}.

In the quantitative linguistics tradition, syllables are considered to be one of the best units (if not the best one) for measuring word length, mainly to warrant cross-linguistic validity \cite{Popescu2013a,Grzybek2013a}. In addition, phonemes (and graphemes) are considered to not be appropriate because they are not immediate constituents of words (words are made of syllables that are made in turn of phonemes). Here we will revise this view arising from research on word lengths and check if it still applies for research on the interaction between dependency distance (in words) and word length.

This article is aimed at testing the hypothesis that DDm implies compression, which we refer to as repertoire unit weight minimization (RUWm). We follow the convention that a lower case $m$ at the end of the abbreviation of a principle indicates "minimization" \cite{Ferrer2019a}.  
RUWm is the minimization of the weight (or cost) of units in a repertoire. In the context of word types and their length as their weight, RUWm corresponds to the minimization of word lengths. RUWm, predicts that more likely units in a repertoire (e.g., more frequent words in a vocabulary) should be lighter (e.g., be shorter). In particular, RUWm has been argued to lead to the emergence of the law of abbreviation across linguistic levels \cite{Ferrer2019c}: e.g., the lexical level where frequent word types or meanings tend to have shorter forms \cite{Zipf1949a,Kanwal2017a,Brochhagen2021a}, and the sublexical level where more frequent cases or case marker types tend to have shorter forms \cite{Liu2021a}.
RUWm is a specialization of a more general principle of unit weight minimization (UWm) arising from research on unifying compression with the origins of both the law of abbreviation and Menzerath's law \cite{Gustison2016a}. The law of abbreviation is the tendency of more frequent words to be shorter \cite{Zipf1949a} while Menzerath's law is the tendency of linguistic constructs with more parts to be made of smaller parts \cite{Altmann1980a}.
A specialization of UWm on sequences, sequence unit weight minimization (SUWm) sheds light on the possible origin of Menzerath's law (\newcite{Gustison2016a}; Table \ref{theory_table}). In the context of words as sequences of syllables and the length of as syllable as its weight (or cost), SUWm corresponds to the minimization of the lengths of syllables in the words they appear. SUWm predicts that units in longer sequences (e.g., syllables in words with more syllables) should have smaller weight (e.g., be shorter). 

Formally, we aim to test   
\begin{equation}
DDm \longrightarrow RUWm. \label{prediction_equation}
\end{equation} 
As an alternative to that prediction one could consider a compensation hypothesis, where less optimized languages at the level of dependency distances would have more pressure for shorter words (or more optimized languages at the level of dependency distances would tolerate longer words). The idea of compensation has been applied in word order research in different ways \cite{Ferrer2015b,Ferrer2014e}. For instance, the suboptimal placement of the verb in SOV orders with respect to DDm (DDm predicts SVO or OVS) has been argued to be compensated by the short length of clitics \cite{Ferrer2014e}.

\begin{sidewaystable}

{\small
\begin{tabular}{lllllll}
{\bf\em Em} & $\longrightarrow$ & {\bf\em Dm} & $\longrightarrow$ & {\bf\em SDm} & $\longrightarrow$   & pairs of primarily alternating word orders \cite{Ferrer2016c} \\
            &                   &             & $\longrightarrow$ & \textcolor{blue}{\bf\em DDm} & $\longrightarrow$ & acceptability \cite{Morrill2000a} \\
            &                   &             &                   &              & $\longrightarrow$   & word order preferences \cite{Morrill2000a} \\
            &                   &             &                   &              & $\longrightarrow$   & scarcity of crossing dependencies \\ 
            &                   &             &                   &              &                     & \cite{Gomez2016a} \\       
            &                   &             &                   &              & $\longrightarrow$   & tendency to uncover the root \cite{Ferrer2008e} \\
            &                   &             &                   &              & $\longrightarrow$   & projectivity \& planarity with high probability \\  
            &                   &             &                   & {\bf\em DDm} (+ projectivity)      & $\longrightarrow$ & medial placement of the root \\
            &                   &             &                   &                                    &                   & \cite{Gildea2007a,Alemany2021a} \\
            &                   &             &                   & {\bf\em DDm} (+ planarity)         & $\longrightarrow$ & medial placement of the central vertex \\ 
            &                   &             &                   &              &                     & \cite{Iordanskii1987a,Hochberg2003a,Alemany2021a} \\            
            &                   &             &                   & {\bf\em DDm} + extreme V placement & $\longrightarrow$ & placement of adjectives with respect to nominal heads \\
            &                   &             &                   & in SVO triples                     &                   & \cite{Ferrer2008e,Ferrer2013e} \\
            &                   &             &                   &                                    & $\longrightarrow$ & placement of auxiliary V with respect to main V \\ 
            &                   &             &                   &                                    &                   & \cite{Ferrer2008e} \\
            &                   &             &                   &                                    & $\longrightarrow$ & consistent branching for dependents of nominal heads \\
            &                   &             &                   &                                    &                   & \cite{Ferrer2014e} \\
            &                   &             &                   &                                    & $\longrightarrow$ & "unnecessity" of headedness parameter \\ 
            &                   &             &                   &                                    &                   & \cite{Ferrer2014e} \\ 
            &                   &             &                   &                                    & $\longrightarrow$ & \textcolor{blue}{\bf\em RUWm} (to be tested in this article) \\             
            & $\longrightarrow$ & {\bf\em UWm} & $\longrightarrow$ & \textcolor{blue}{\bf\em RUWm} & $\longrightarrow$  & Zipf's law of abbreviation \\
            &                   &            &                   &                            &                     & \cite{Shannon1948,Ferrer2019c} \\
            &                   &            &                   &                            &                     & reduction \\ 
            &                   &            &                   &                            &                     & \cite{Ferrer2013f} \\    
            &                   &            &                   & \textcolor{blue}{\bf\em RUWm} + & $\longrightarrow$ & {\bf\em Hm} \cite{Ferrer2015b} \\
            &                   &            &                   & unique segmentation \\
            &                   &            &                   & {\bf\em SUWm}                   & $\longrightarrow$  & Menzerath's law \\ 
            &                   &            &                   &                                &                     & \cite{Gustison2016a,Ferrer2019c} \\            
\end{tabular}
}
\caption{\label{theory_table} Optimization principles and their predictions. Arrows link principles with their predictions. Predictions can take the form of principles or manifestations. Principles are marked in boldface.
The two principles whose relationship is the target of this article are marked in blue. {\em Em}: energy minimization. {\em Dm}: distance minimization. {\em UWm}: unit weight minimization, popularly known as {\em compression}. {\em RUWm}: repertoire unit weight minimization. {\em SUWm}: sequence unit weight minimization. {\em SDm}: swap distance minimization. {\em DDm}: dependency distance minimization. {\em Hm}: entropy (or surprisal) minimization. We use parentheses for assumptions that are likely to be predictions of DDm and thus likely to be unnecessary assumptions to a large extent (see \protect \newcite{Gomez2016a} for further details about the argument). Awareness of such unnecessary assumptions is vital for the construction of a parsimonious theory. } 
\end{sidewaystable}

\section{Methods}

To measure dependency distance in a sentence, we considered two scores, both computed using dependency distances measured in words. The first score is $\Omega$, a recently introduced normalized score that takes the value $1$ when the dependency distances are fully optimized and is expected to take a value of $0$ if there is no bias on dependency distances, for or against DDm \cite{Ferrer2020b}. $\Omega$ can be seen as score of the intensity of DDm, which is maximum when $\Omega=1$, and missing when $\Omega\approx 0$. $\Omega<0$ indicates that DDm is surpassed by other word order principles. The score has the virtue of satisfying a series of mathematical and statistical properties: dual normalization (i.e. normalization with respect to both the minimum and the random baseline), constancy under minimum linear arrangement, stability under random linear arrangement, invariance under linear transformation and boundedness under maximum linear arrangement \cite{Ferrer2020b}. These properties are particularly useful when calculating an average score over the sentences of a treebank. For comparison, we consider also $D$, the sum of dependencies of a sentence.
While $\Omega$ is a measure of closeness, $D$ is a measure of distance. $D$ is the most widely used score \cite{Gildea2007a,Gildea2010a,Futrell2015a,Futrell2020a} but does not satisfy any of the remarkable mathematical properties enumerated above. The intensity of DDm is difficult to assess just from the value of $D$.   

For each treebank considered in this study, we calculated an average $\Omega$ and average $D$ over all the sentences of the treebank.  
To measure word length, we considered two different units: phonemes and syllables. For each language, we calculated $L_s$, the mean word length in syllables (mean syllables per word token) and $L_p$, the mean word length in phonemes (mean phonemes per word token).

To control for the content or the source text of the treebanks, we used a parallel collection of treebanks, in particular, the Parallel Universal Dependencies (PUD) collection version 2.6 \cite{conll2017st_APS}. PUD contains 20 languages from 9 distinct families. PUD follows the UD annotation style \cite{ud26_APS}. To control for annotation style, we also use SUD, i.e. Surface-Syntactic Universal Dependencies \cite{sud}. We use PSUD to refer to the PUD collection following the SUD annotation \cite{Ferrer2020b}. The PUD and PSUD treebank collections are borrowed from a recent study \cite{Ferrer2020b}. The preprocessing of these treebanks involves the removal of punctuation marks and reparalellization to warrant there is no loss of parallelism after punctuation mark removal \cite{Ferrer2020b}. The data are available from \url{https://github.com/lluisalemanypuig/optimality-syntactic-dependency-distances} in two levels of preprocessing: (a) the preprocessed treebanks as head vectors and (b) the raw text tables that were extracted from them and used to feed the statistical analyses. The transformation of the raw head vectors into the raw text tables can be replicated easily with the Linear Arrangement Library \cite{Alemany2022a}. 
  
The PUD data does not include syllable or phoneme annotations that would allow us to measure word lengths in these units, and obtaining them would be highly costly. Thus, we instead borrowed mean word lengths
from the dataset of another recent study \cite{Fenk-Oczlon2021a}. In that study, mean word lengths were estimated from 22 simple declarative sentences encoding one proposition and using basic vocabulary.
Three languages from PUD/PSUD are missing in that dataset: Arabic (Afro-Asiatic), Indonesian (Austronesian) and Swedish (Indo-European). As a result, the final collection has 17 languages from 7 distinct families that are displayed in Table \ref{typological_diversity_table}. 

We consider two approaches to investigate the relationship between dependency distance and word length. First, a Kendall $\tau$ correlation test between the mean dependency distance score ($D$ or $\Omega$) and mean word length ($L_s$ or $L_p$). With respect to plain Pearson correlation, $\tau$ is more robust to extreme observations and to non-linearity \cite{Newson2002}. A significant negative correlation between mean $\Omega$ and $L_s$ or $L_p$ would confirm the prediction in Eq. \ref{prediction_equation}. 
Note that $\Omega$ and $D$ have opposite interpretations (larger values of $\Omega$ imply shorter dependencies whereas larger values of $D$ imply longer dependencies) hence a positive correlation with respect to $D$ would be analogous to a negative correlation with respect to $\Omega$. Therefore, a positive correlation between $D$ and $L_s$ or $L_p$ could also be interpreted as confirming the prediction in Eq. \ref{prediction_equation} but $D$ lacks the mathematical and statistical properties that are required for robust assessment \cite{Ferrer2020b}.

Second, the fact that the Indo-European family is over-represented and the only one that is represented by more than one language (Table \ref{typological_diversity_table}) motivates the need to control for the effect of family size. Accordingly, we also consider a couple of kinds of  generalized linear models. First, a null model with mean word length as response ($L_s$ or $L_p$) and language family as random factor. Second, a mixed effects model with mean word length as response ($L_s$ or $L_p$), language family as random effect and a dependency distance score (mean $D$ or mean $\Omega$) as fixed effect. To test the prediction with these models, we use information theoretic model selection \cite{Burnham2002a,Winter2019a}. 
AIC of the mixed effects model being lower than that of the null model would confirm the prediction provided that the weight of the association between the predictor and the response 
has a sign that matches that of the prediction. 
Again, some caution is needed when $D$ is involved because of its technical limitations \cite{Ferrer2020b}.

The linear models were fitted with the {\tt lme4} R package. Confidence intervals for the weight of the fixed effect were computed using the parametric bootstrapping method of function {\tt confint} of the {\tt lme4} package.
  
\begin{table*}
\centering
\begin{tabular}{lp{4in}}
Family & Languages \\
\hline
Turkic (1) & Turkish \\
Indo-European (11) & Czech, English, French, German, Hindi, Icelandic, Italian, Polish, Portuguese, Russian, Spanish \\ 
Japonic (1) & Japanese \\
Koreanic (1) & Korean \\
Sino-Tibetan (1) & Chinese \\
Tai-Kadai (1) & Thai \\
Uralic (1) & Finnish \\
\end{tabular}
\caption[Typological diversity of the treebank collections used in this study.]{\label{typological_diversity_table} The 17 languages from 7 families used in the present study. The counts attached to each family name indicate the number of different languages included in the present study. }
\end{table*}

\section{Results}

\begin{figure}[tbp]
\begin{center}
\includegraphics[width = \textwidth]{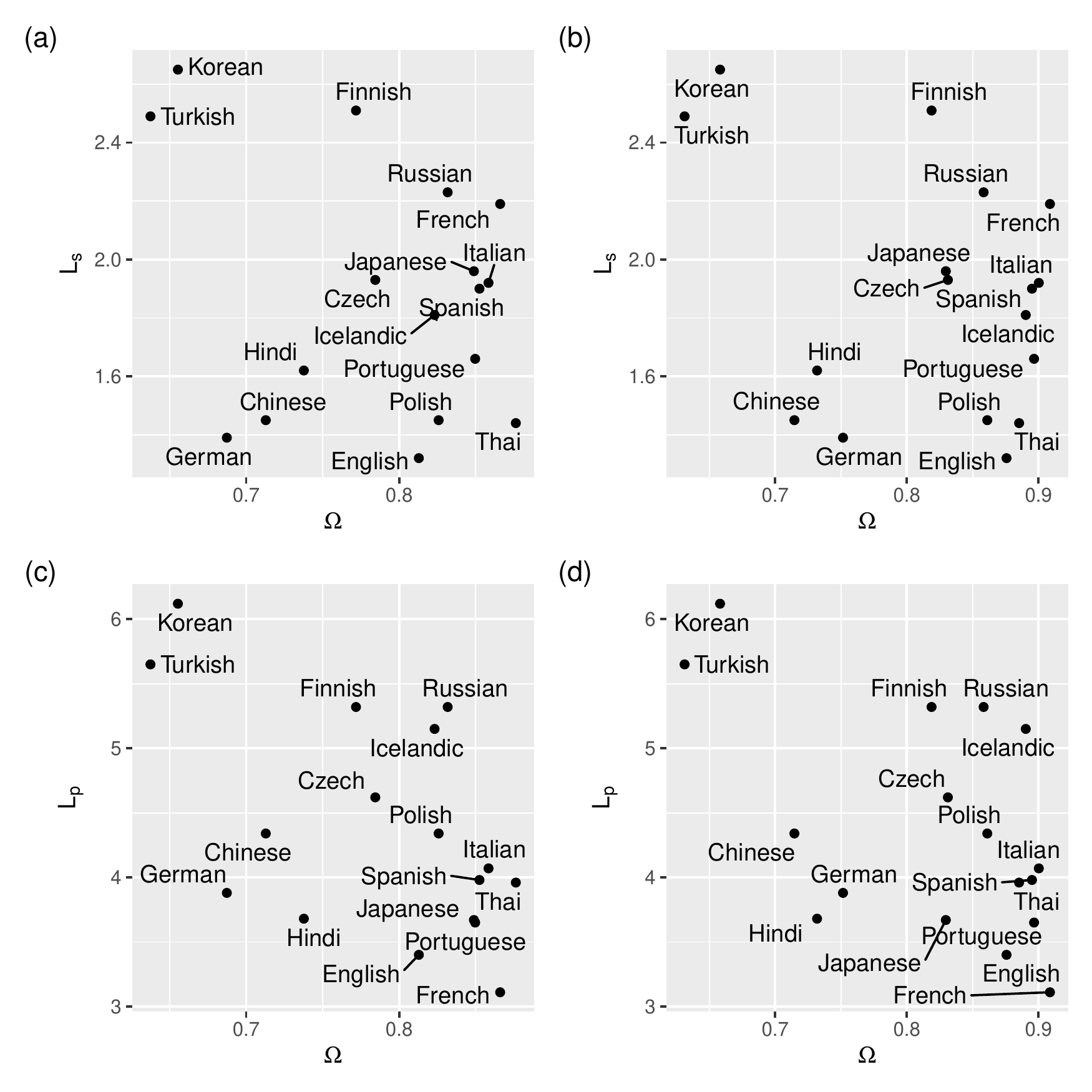}
\end{center}
\caption{\label{mean_word_length_versus_Omega_figure} Mean word length ($L$) as a function of the mean degree of optimality of syntactic dependency distances (mean $\Omega$). (a) $L_s$ as a function of mean $\Omega$ in PUD. (b) $L_s$ as a function of mean $\Omega$ in PSUD. (c) $L_p$ as a function of mean $\Omega$ in PUD. (d) $L_p$ as a function of mean $\Omega$ in PSUD.  }
\end{figure}

\begin{figure}[tbp]
\begin{center}
\includegraphics[width = \textwidth]{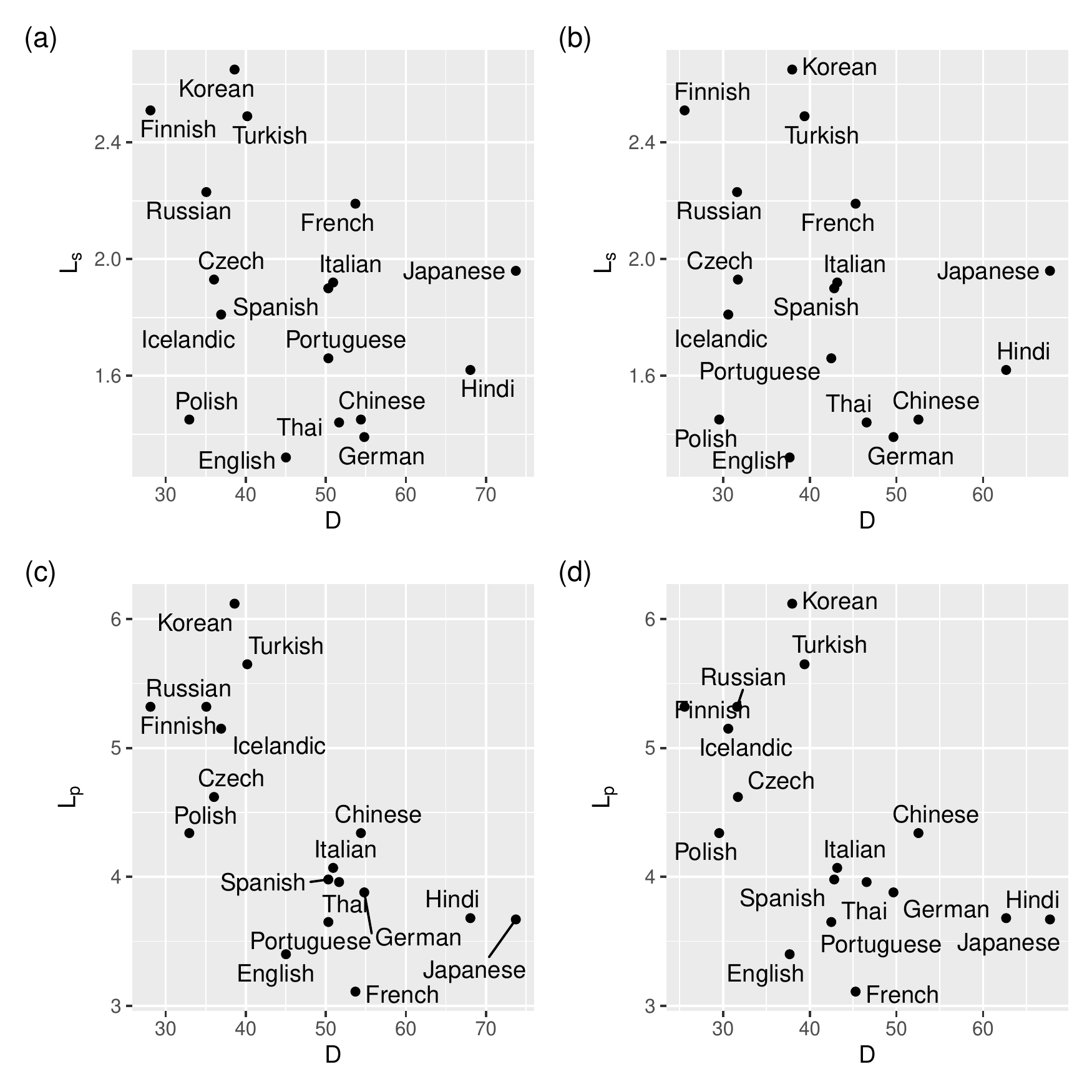}
\end{center}
\caption{\label{mean_word_length_versus_D_figure} Mean word length ($L$) as a function of the mean sum of dependency distances (mean $D$). (a) $L_s$ as a function of $D$ in PUD. (b) $L_s$ as a function of $D$ in PSUD. (c) $L_p$ as a function of mean $D$ in PUD. (d) $L_p$ as a function of mean $D$ in PSUD.  }
\end{figure}

\begin{table}[tbp]
\begin{center}
\begin{tabular}{lllllll}
Collection & distance & length & $n$ & $\tau$ & $p$ \\
\hline 
PUD & $\Omega$ & $L_s$ & 17 & -0.052 & 0.773 \\
& & $L_p$ & 17 & -0.37 & 0.039 \\
PSUD & $\Omega$ & $L_s$ & 17 & -0.111 & 0.536 \\
& & $L_p$ & 17 & -0.37 & 0.039 \\
PUD & $D$ & $L_s$ & 17 & -0.258 & 0.149 \\
 & & $L_p$ & 17 & -0.459 & 0.011 \\
PSUD & $D$ & $L_s$ & 17 & -0.185 & 0.303 \\
 & & $L_p$ & 17 & -0.385 & 0.032 \\

\end{tabular}
\end{center}
\caption{\label{correlations_table} The correlation between the mean dependency distance score and mean word length. We show the annotation style, the distance score, the word length score, the value of the Kendall $\tau$ correlation statistic and $p$, the p-value of the corresponding two-sided test. We assume a significance level of 0.05. } 
\end{table}

\begin{table}[tbp]
\begin{center}
\begin{tabular}{lllll}
Collection & distance & length & AIC mixed effects & AIC null \\
\hline
PUD & $\Omega$ & $L_s$ & 23.15 & 23.9 \\
 & & $L_p$ & 42.16 & 47.64 \\
PSUD & $\Omega$ & $L_s$ & 23.54 & 23.9 \\
 & & $L_p$ & 42.96 & 47.64 \\
PUD & $D$ & $L_s$ & 32.28 & 23.9 \\
 & & $L_p$ & 45.23 & 47.64 \\
PSUD & $D$ & $L_s$ & 32.22 & 23.9 \\
 & & $L_p$ & 48.03 & 47.64 \\
 
\end{tabular}
\end{center}
\caption{\label{AIC_table} Information theoretic selection of models to predict the mean word length. We show the annotation style, the distance score, the word length score, the Akaike Information Criterion (AIC) of the mixed effects model and the AIC of the null model.} 
\end{table}

Fig. \ref{mean_word_length_versus_Omega_figure} shows the relationship between word length and $\Omega$. Table \ref{correlations_table} indicates that the correlation between $\Omega$ and $L_p$ is negative and statistically significant whereas the correlation between $\Omega$ and $L_s$ is also negative but not significant. That is, the shorter the syntactic dependencies of a language upon dual normalization ($\Omega$), the shorter the words when their length is measured in phonemes. 
Table \ref{AIC_table} indicates that the result is confirmed by a linear mixed effects model that predicts $L_s$ or $L_p$ based on $\Omega$ with family as random effect. The null model (only family as random factor) always yields an AIC value that is larger than that of the mixed effects model ($\Omega$ as fixed effect and family as random effect). However, the AIC of the null model is only sufficiently large (i.e., differing by various units) when the predictor is $L_s$, in agreement with the plain correlation analysis (Table \ref{correlations_table}). 
The analysis of the confidence intervals supports the conclusions: the confidence interval for the weight of $\Omega$ with $L_p$ as response comprises exclusively negative values, consistently with a negative correlation between $\Omega$ and $L_p$ (Table \ref{confidence_interval_table}).

\begin{table}[tbp]
\begin{center}
\begin{tabular}{lllll}
Collection & Fixed effect & Response & Lower & Upper \\
\hline
PUD & $\Omega$ & $L_p$ & -10.85 & -1.19 \\
PSUD & $\Omega$ & $L_p$ & -9.21 & -0.69 \\
PUD & $D$ & $L_s$ & -0.03 & 0 \\
 & & $L_p$ & -0.07 & -0.02 \\
PSUD & $D$ & $L_s$ & -0.03 & 0.01 \\
 & & $L_p$ & -0.07 & -0.02 \\
 
\end{tabular}
\end{center}
\caption{\label{confidence_interval_table} Lower and upper bounds of the $95\%$ confidence interval for the weight of the fixed effect ($\Omega$ or $D$) in the mixed effects models. The confidence intervals for $\Omega$ with $L_p$ as response could not be computed due to numerical problems. } 
\end{table}

If $\Omega$ is replaced by $D$ in the preceding analyses, opposite conclusions are drawn concerning the direction of the correlation. First, Figure \ref{mean_word_length_versus_D_figure} suggests that mean word length tends to decrease as $D$ increases when word length is measured in phonemes in both PUD and PSUD. 
Accordingly,
Table \ref{correlations_table} confirms that the correlation between $D$ and $L_p$ is negative and statistically significant while the correlation between $D$ and $L_s$ turns out to not be significant. That is, the longer the syntactic dependencies of a language, the shorter the words when their length is measured in phonemes, the opposite conclusion that is reached with $\Omega$. Table \ref{AIC_table} indicates that the results are confirmed by linear mixed effects linear models that predict $L_s$ or $L_p$ based on $D$ with family as random effect. When measuring word lengths in syllables, the AIC of the null model is much smaller than that of the mixed effects model with $D$ as predictor, supporting that $D$ and $L_s$ are uncorrelated. In contrast, the AIC of the null model is larger (by more than unit) than that of the mixed effects model with $D$ as predictor, supporting that $D$ and $L_p$ are correlated. 
The analysis of the confidence intervals (Table \ref{confidence_interval_table}) confirms a negative correlation between $D$ and $L_p$ because the confidence intervals for the weight of $D$ with $L_p$ as response comprise exclusively negative values; that does not happen when $L_s$ is the response.

\section{Discussion}


We have confirmed the prediction that DDm leads to compression (Eq. \ref{prediction_equation}) by showing that $\Omega$ and $L_p$ are negatively correlated: the closer the syntactically related words, the shorter the words. Although syllables are the preferred unit of measurement of word length over phonemes in quantitative linguistics \cite{Popescu2013a,Grzybek2013a}, we have failed to confirm the prediction with that unit. 
We cannot exclude the possibility that a negative correlation between $\Omega$ and $L_s$ exists but has not surfaced because it is weaker and the sample of languages is not large enough. However, the failure with syllables may be a confirmation of the higher capacity of phonemes over syllables to capture the so-called 'phonological complexity' \cite{Pimentel2020a}. Furthermore, we suspect that word lengths in phonemes lead eventually to a more accurate estimation of the true distance between words, currently measured in words, than syllables. To see the relationship between word length and dependency distance and to build an explanation, suppose that $\delta_p$ is the phonemic distance between two words, $\delta_s$ is their syllabic distance and $d$ is their distance in words ($d=1$ if two words are consecutive). Then assuming that a word and its dependent word are connected by the middle of their phonemic or syllabic sequence, a mean field approximation yields $\delta_p \approx d L_p$ (the head and their dependent contribute with $L_p/2$ phonemes; the words in-between the head and the governor contribute with $(d-1)L_p$ phonemes). Analogously, $\delta_s \approx d L_s$. Then $\delta_p$ may be more strongly correlated with the time that is needed to keep unresolved dependencies in memory \cite{Morrill2000a} than $\delta_s$ because syllables vary in phonemic length. That time was considered to be the key to understand what determines the acceptability of sentences and other word order phenomena \cite{Morrill2000a}. Then $\delta_p$ looks like a better proxy for the combination of memory decay and interference that is believed to cause DDm \cite{Liu2017a,Temperley2018a}. To recap, the argument that syllables are more appropriate units than phonemes (or graphemes) because syllables are immediate word constituents \cite{Popescu2013a,Grzybek2013a} may not be appropriate for research on dependency distances because distances (or time) could be measured more accurately with constituents located farther down in the hierarchy of constituents.

We have also shown that $D$, a widely used dependency score \cite{Gildea2007a,Gildea2010a,Futrell2015a,Futrell2020a}, fails to confirm the prediction that DDm implies compression. It is not the first time that $\Omega$ shows a superior performance when testing theoretical predictions. When testing that DDm should be surpassed by other word order principles in short sentences \cite{Ferrer2014a,Ferrer2013f}, $\Omega$ was able to find many more languages with anti-DDm effects than $D$ \cite{Ferrer2020b}. Our findings reinforce the view that raw dependency distances are a poor reflect of DDm and that advanced scores such as $\Omega$ are crucial for progress in research on that optimization principle and related memory constraints \cite{Ferrer2020b}, and eventually, for the construction of a general theory of natural systems with human language as a particular case and energy minimization at the center \cite{Semple2021a}. An early sketch of that theory is shown in Table \ref{theory_table}.

For each language, we have measured dependency distances ($\Omega$ and $D$) on the collection of sentences included in PUD/PSUD whereas mean word lengths ($L_p$ and $L_s$) come from an independent collection of sentences \cite{Fenk-Oczlon2021a}. It could be argued that mean word lengths should have been estimated on PUD/PSUD, too. 
This would make the whole study fully parallel, and word length data potentially more accurate, as it would come from a larger sample and, more specifically, from the same sentences where dependency lengths have been measured. It is conceivable, for example, that a given kind of syntactic construction could produce shorter dependency distances in language A than in language B, while being prevalent in PUD/PSUD but not in the word length dataset (e.g. due to genre, style or topic differences). This could cause us to observe increased dependency distance minimization in language A with respect to B without being able to observe the associated compression, as our word length data would not include sentences with that specific phenomenon. In turn, this could cause us to underestimate the correlation between $\Omega$ and $L_s$ or $L_p$.
Unfortunately,  the technical complexity of replicating the present study with actual phonemic or syllabic lengths on PUD/PSUD goes beyond the scope of the present article. 
These considerations notwithstanding, notice that the stress of this article is on testing a prediction rather than theoretically agnostic data description. In this context, it is rather astonishing that the theoretical prediction (Eq. \ref{prediction_equation}) is confirmed even though the collections of sentences for dependency distances and the collections for word lengths are independent. That confirmation offers two major interpretations: (a) the results are due to biases in the sentences, either in PUD or in Fenk-Oczlon's dataset \cite{Fenk-Oczlon2021a} or, crucially, (b) there is actually a deep reason for the prediction to hold. Further research on a fully parallel scenario or alternative sources is required.
 
By having shown that DDm implies compression of word lengths, we do not mean that compression is produced exclusively by DDm. 
Our work does not rule out other sources for compression. Following previous research testing successfully the prediction that DDm weakens in small sequences \cite{Ferrer2019a,Ferrer2020b}, we formulate a new prediction, namely that compression {\em per se} (independently from DDm) may surface in short sequences. Future research should clarify the weight of the contribution to compression from DDm, compression itself and other principles.  

Once one integrates our findings into the piece of a mathematical theory of communication in Table \ref{theory_table}, it turns out that we have actually uncovered the following chain,
\begin{eqnarray*}
Dm \longrightarrow DDm \longrightarrow RUWm \longrightarrow & \mbox{Zipf's law of abbreviation} 
\end{eqnarray*}
Namely, a general principle of minimization of the distance between elements leads to the prediction of the law of abbreviation. 

Our findings have implications for competing views and frameworks. 
First, the compensation hypothesis outlined in the introduction, i.e. less optimized languages at the level of dependency distances would have more pressure for shorter words (or more optimized languages at the level of dependency distances would tolerate longer words), would predict that the correlation between $\Omega$ and word length should not be negative. Instead, a zero or positive correlation would be expected depending on the strength or the nature of the compensation effect. Our findings of a negative correlation rule out the compensation hypothesis as a primary explanation for the global trend. However, we cannot exclude that compensation has some secondary role in general, an important role in specific languages or an important role with certain domains of a language, as suggested for the suboptimal placement of clitics with respect to DDm in Romance languages \cite{Ferrer2014e}.	
Second, DDm predicts compression and compression in turn predicts reduction (see Section 3.4 of \newcite{Ferrer2013f}). By reduction, here we mean  the shortening or omission of predictable utterances, a phenomenon that has  been  used  to justify the uniform information density and related hypotheses (see \newcite{Lemke2021a} and references therein). Therefore, the need for uniform  information  density  and related hypotheses as standalone hypotheses needs to be revised. Third, our finding of an association between word length and dependency distance also challenges the recent conclusion that {\em (compositional) morphology and graphotactics can sufficiently account for most of the complexity of natural codes -- as measured by code length} \cite{Pimentel2021a} as if no strong independent pressure for compression (beyond morphology and graphotactics) really existed. Our findings suggest that the two components of the problem of compression as defined in standard information theory, i.e. the minimization of word length and the conditions of the coding scheme \cite{Ferrer2019c}, may not be as easy to dissociate from (compositional) morphology and graphotactics as expected from the traditional reductionistic division into linguistic levels. Indeed, our article demonstrates how constraints on the ``syntactic level'' (DDm) may be shaping the ``(sub)lexical level'' (compression on word lengths). 

Our work has implications beyond the current state of development of the theory of the communication efficiency reviewed in Table \ref{theory_table}. We have confirmed that DDm implies compression but the outcome of our correlation analysis does not exclude that it could also the other way around, namely that compression is actually leading to DDm. A possible track could be that pressure for shorter words may lead to a loss of information that would require more words to convey the same message, which in turn would imply longer sentences and then higher pressure for DDm. Another track could be that a general principle of compression operates on top both at the level of words and at higher levels (phrases, clauses, sentences) and then close packaging (DDm) at all these levels is simply a consequence of maximizing compression, in line with the now-or-never bottleneck \cite{Christiansen2015a}. We hope that our work stimulates further research on general principles and their predictions.

\iftoggle{anonymous}{}
{ 
\section*{Acknowledgements}

We are grateful to S. Semple and M. Gustison for extremely valuable and inspiring discussions on theory construction \cite{Semple2021a}. 
We are also grateful to two anonymous reviewers, T. Pimentel and G. Fenk-Oczlon for helpful comments. 
RFC is supported by the grant TIN2017-89244-R from MINECO (Ministerio de Econom\'ia, Industria y Competitividad) and the recognition 2017SGR-856 (MACDA) from AGAUR (Generalitat de Catalunya). 
CGR is supported by the European Research Council (ERC), under the European Union's Horizon 2020 research and innovation
programme (FASTPARSE, grant agreement No 714150), the ANSWER-ASAP (TIN2017-85160-C2-1-R) and SCANNER-UDC (PID2020-113230RB-C21) projects from ERDF/MICINN/AEI, 
Xunta de Galicia (ED431C 2020/11 and an Oportunius program grant to complement ERC grants); the CITIC research center is funded by ERDF and Xunta de Galicia (ERDF - Galicia 2014-2020 program, grant ED431G 2019/01).
}

\bibliographystyle{acl}
\bibliography{../biblio_dt/main,../biblio_dt/twoplanaracl,../biblio_dt/Ramon,../biblio_dt/optimization_in_biology,../biblio_dt/twoplanaracl_ours,../biblio_dt/Ramon_ours}

\newcommand{\beeksort}[1]{}
\begin{thebibliography}{}

\bibitem[\protect\citename{{Alemany-Puig} and
  {Ferrer-i-Cancho}}2022]{Alemany2022a}
Llu\'is {Alemany-Puig} and Ramon {Ferrer-i-Cancho}.
\newblock 2022.
\newblock The {Linear Arrangement Library. A} new tool for research on
  syntactic dependency structures.
\newblock In {\em Proceedings of the SyntaxFest 2022}, Sofia, Bulgaria.
  Association for Computational Linguistics.

\bibitem[\protect\citename{{Alemany-Puig} \bgroup et al.\egroup
  }2022]{Alemany2021a}
Llu\'is {Alemany-Puig}, Juan~Luis Esteban, and Ramon {Ferrer-i-Cancho}.
\newblock 2022.
\newblock Minimum projective linearizations of trees in linear time.
\newblock {\em Information Processing Letters}, 174:106204.

\bibitem[\protect\citename{Altmann}1980]{Altmann1980a}
Gabriel Altmann.
\newblock 1980.
\newblock Prolegomena to {Menzerath's} law.
\newblock {\em Glottometrika}, 2:1--10.

\bibitem[\protect\citename{Behaghel}1932]{Behaghel1932a}
Otto Behaghel, 1932.
\newblock {\em Deutsche Syntax: eine geschichtliche Darstellung}, chapter Bd 4:
  Wortstellung. Periodenbau.
\newblock Carl Winters Univeritätsbuchhandlung, Heidelberg, Germany.

\bibitem[\protect\citename{Brochhagen}2021]{Brochhagen2021a}
Thomas Brochhagen.
\newblock 2021.
\newblock Brief at the risk of being misunderstood: Consolidating
  population-and individual-level tendencies.
\newblock {\em Computational Brain {\&} Behavior}, Feb.

\bibitem[\protect\citename{Burnham and Anderson}2002]{Burnham2002a}
Kenneth~P. Burnham and David~R. Anderson.
\newblock 2002.
\newblock {\em Model selection and multimodel inference. {A} practical
  information-theoretic approach}.
\newblock Springer, New York, 2nd edition.

\bibitem[\protect\citename{Christiansen and Chater}2016]{Christiansen2015a}
Morten~H. Christiansen and Nick Chater.
\newblock 2016.
\newblock The now-or-never bottleneck: a fundamental constraint on language.
\newblock {\em Behavioral and Brain Sciences}, 39:1 -- 72.

\bibitem[\protect\citename{Fenk-Oczlon and Pilz}2021]{Fenk-Oczlon2021a}
Gertraud Fenk-Oczlon and Jürgen Pilz.
\newblock 2021.
\newblock Linguistic complexity: Relationships between phoneme inventory size,
  syllable complexity, word and clause length, and population size.
\newblock {\em Frontiers in Communication}, 6:66.

\bibitem[\protect\citename{{Ferrer-i-Cancho} and
  {G\'omez-Rodr\'iguez}}2021]{Ferrer2019a}
Ramon {Ferrer-i-Cancho} and Carlos {G\'omez-Rodr\'iguez}.
\newblock 2021.
\newblock Anti dependency distance minimization in short sequences. a graph
  theoretic approach.
\newblock {\em Journal of Quantitative Linguistics}, 28(1):50--76.

\bibitem[\protect\citename{{Ferrer-i-Cancho} \bgroup et al.\egroup
  }2019]{Ferrer2019c}
Ramon {Ferrer-i-Cancho}, Christian Bentz, and Caio Seguin.
\newblock 2019.
\newblock Optimal coding and the origins of {Zipfian} laws.
\newblock {\em Journal of Quantitative Linguistics}, page in press.

\bibitem[\protect\citename{{Ferrer-i-Cancho} \bgroup et al.\egroup
  }2021]{Ferrer2020b}
Ramon {Ferrer-i-Cancho}, Carlos {G\'omez-Rodr\'iguez}, Juan~Luis Esteban, and
  Llu\'is {Alemany-Puig}.
\newblock 2021.
\newblock The optimality of syntactic dependency distances.
\newblock {\em Physical Review E}, page in press.

\bibitem[\protect\citename{{Ferrer-i-Cancho}}2003]{Ferrer2003i}
Ramon {Ferrer-i-Cancho}.
\newblock 2003.
\newblock {\em Language: universals, principles and origins}.
\newblock {Ph.D.} thesis, Universitat Polit\`ecnica de Catalunya, Barcelona.

\bibitem[\protect\citename{{Ferrer-i-Cancho}}2004]{Ferrer2004b}
Ramon {Ferrer-i-Cancho}.
\newblock 2004.
\newblock {Euclidean} distance between syntactically linked words.
\newblock {\em Physical Review E}, 70:056135.

\bibitem[\protect\citename{{Ferrer-i-Cancho}}2008]{Ferrer2008e}
Ramon {Ferrer-i-Cancho}.
\newblock 2008.
\newblock Some word order biases from limited brain resources. {A} mathematical
  approach.
\newblock {\em Advances in Complex Systems}, 11(3):393--414.

\bibitem[\protect\citename{{Ferrer-i-Cancho}}2014]{Ferrer2014a}
Ramon {Ferrer-i-Cancho}.
\newblock 2014.
\newblock Why might {SOV} be initially preferred and then lost or recovered?
  {A} theoretical framework.
\newblock In E.~A. Cartmill, S.~Roberts, H.~Lyn, and H.~Cornish, editors, {\em
  {The Evolution of Language - Proceedings of the 10th International Conference
  (EVOLANG10)}}, pages 66--73, Vienna, Austria. Wiley.
\newblock {Evolution} of {Language} {Conference} ({Evolang} 2014), April 14-17.

\bibitem[\protect\citename{{Ferrer-i-Cancho}}2015a]{Ferrer2013e}
Ramon {Ferrer-i-Cancho}.
\newblock 2015a.
\newblock The placement of the head that minimizes online memory. {A} complex
  systems approach.
\newblock {\em Language Dynamics and Change}, 5(1):114--137.

\bibitem[\protect\citename{{Ferrer-i-Cancho}}2015b]{Ferrer2014e}
Ramon {Ferrer-i-Cancho}.
\newblock 2015b.
\newblock Reply to the commentary ``{Be careful when assuming the obvious}'',
  by {P. Alday}.
\newblock {\em Language Dynamics and Change}, 5(1):147--155.

\bibitem[\protect\citename{{Ferrer-i-Cancho}}2016]{Ferrer2016c}
Ramon {Ferrer-i-Cancho}.
\newblock 2016.
\newblock Kauffman's adjacent possible in word order evolution.
\newblock In {\em The evolution of language: proceedings of the 11th
  International Conference (EVOLANG11)}.

\bibitem[\protect\citename{{Ferrer-i-Cancho}}2017a]{Ferrer2013f}
Ramon {Ferrer-i-Cancho}.
\newblock 2017a.
\newblock The placement of the head that maximizes predictability. {An}
  information theoretic approach.
\newblock {\em Glottometrics}, 39:38--71.

\bibitem[\protect\citename{{Ferrer-i-Cancho}}2017b]{Ferrer2017c}
Ramon {Ferrer-i-Cancho}.
\newblock 2017b.
\newblock Towards a theory of word order. comment on "dependency distance: {A}
  new perspective on syntactic patterns in natural language" by {Haitao Liu} et
  al.
\newblock {\em Physics of Life Reviews}, 21:218 -- 220.

\bibitem[\protect\citename{{Ferrer-i-Cancho}}2018]{Ferrer2015b}
Ramon {Ferrer-i-Cancho}.
\newblock 2018.
\newblock Optimization models of natural communication.
\newblock {\em Journal of Quantitative Linguistics}, 25(3):207--237.

\bibitem[\protect\citename{Futrell \bgroup et al.\egroup }2015]{Futrell2015a}
Richar Futrell, Kyle Mahowald, and Edward Gibson.
\newblock 2015.
\newblock Large-scale evidence of dependency length minimization in 37
  languages.
\newblock {\em Proceedings of the National Academy of Sciences},
  112(33):10336--10341.

\bibitem[\protect\citename{Futrell \bgroup et al.\egroup }2020]{Futrell2020a}
Richard Futrell, Roger~P. Levy, and Edward Gibson.
\newblock 2020.
\newblock Dependency locality as an explanatory principle for word order.
\newblock {\em Language}, 96(2):371--412.

\bibitem[\protect\citename{Gerdes \bgroup et al.\egroup }2018]{sud}
Kim Gerdes, Bruno Guillaume, Sylvain Kahane, and Guy Perrier.
\newblock 2018.
\newblock {SUD} or surface-syntactic universal dependencies: An annotation
  scheme near-isomorphic to {UD}.
\newblock In {\em Proceedings of the Second Workshop on Universal Dependencies
  ({UDW} 2018)}, pages 66--74, Brussels, Belgium, November. Association for
  Computational Linguistics.

\bibitem[\protect\citename{Gildea and Temperley}2007]{Gildea2007a}
Daniel Gildea and David Temperley.
\newblock 2007.
\newblock Optimizing grammars for minimum dependency length.
\newblock In {\em Proceedings of the 45th Annual Meeting of the Association of
  Computational Linguistics}, pages 184--191, Prague, Czech Republic, June.
  Association for Computational Linguistics.

\bibitem[\protect\citename{Gildea and Temperley}2010]{Gildea2010a}
Daniel Gildea and David Temperley.
\newblock 2010.
\newblock Do grammars minimize dependency length?
\newblock {\em Cognitive Science}, 34(2):286--310.

\bibitem[\protect\citename{{G\'omez-Rodr\'iguez} and
  {Ferrer-i-Cancho}}2017]{Gomez2016a}
Carlos {G\'omez-Rodr\'iguez} and Ramon {Ferrer-i-Cancho}.
\newblock 2017.
\newblock Scarcity of crossing dependencies: {A} direct outcome of a specific
  constraint?
\newblock {\em Physical Review E}, 96:062304.

\bibitem[\protect\citename{Grzybek}2013]{Grzybek2013a}
Peter Grzybek.
\newblock 2013.
\newblock Word length.
\newblock In John~R. Taylor, editor, {\em The Oxford Handbook of the Word}.
  Oxford University Press.

\bibitem[\protect\citename{Gustison \bgroup et al.\egroup }2016]{Gustison2016a}
Morgan~L. Gustison, Stuart Semple, Rarmon {Ferrer-i-Cancho}, and Thore Bergman.
\newblock 2016.
\newblock Gelada vocal sequences follow {Menzerath}'s linguistic law.
\newblock {\em Proceedings of the National Academy of Sciences USA},
  13:E2750--E2758.

\bibitem[\protect\citename{Hochberg and Stallmann}2003]{Hochberg2003a}
Robert~A. Hochberg and Matthias~F. Stallmann.
\newblock 2003.
\newblock Optimal one-page tree embeddings in linear time.
\newblock {\em Information Processing Letters}, 87:59--66.

\bibitem[\protect\citename{Iordanskii}1987]{Iordanskii1987a}
Mikhail~A. Iordanskii.
\newblock 1987.
\newblock Minimal numberings of the vertices of trees --- {Approximate}
  approach.
\newblock In Lothar Budach, Rais~Gati{\v{c}} Bukharajev, and
  Oleg~Borisovi{\v{c}} Lupanov, editors, {\em Fundamentals of Computation
  Theory}, pages 214--217, Berlin, Heidelberg. Springer Berlin Heidelberg.

\bibitem[\protect\citename{Kanwal \bgroup et al.\egroup }2017]{Kanwal2017a}
Jasmeen Kanwal, Kenny Smith, Jennifer Culbertson, and Simon Kirby.
\newblock 2017.
\newblock Zipf's law of abbreviation and the principle of least effort:
  Language users optimise a miniature lexicon for efficient communication.
\newblock {\em Cognition}, 165:45--52.

\bibitem[\protect\citename{Lemke \bgroup et al.\egroup }2021]{Lemke2021a}
Robin Lemke, Lisa Schäfer, and Ingo Reich.
\newblock 2021.
\newblock Modeling the predictive potential of extralinguistic context with
  script knowledge: The case of fragments.
\newblock {\em PLoS One}, 16:e0246255.

\bibitem[\protect\citename{Liu \bgroup et al.\egroup }2017]{Liu2017a}
Haitao Liu, Chunshan Xu, and Junying Liang.
\newblock 2017.
\newblock Dependency distance: {A} new perspective on syntactic patterns in
  natural languages.
\newblock {\em Physics of Life Reviews}, 21:171--193.

\bibitem[\protect\citename{Liu}2008]{Liu2008a}
Haitao Liu.
\newblock 2008.
\newblock Dependency distance as a metric of language comprehension difficulty.
\newblock {\em Journal of Cognitive Science}, 9:159--191.

\bibitem[\protect\citename{Liu}2021]{Liu2021a}
Zoey Liu.
\newblock 2021.
\newblock A multifactorial approach to crosslinguistic constituent orderings.
\newblock {\em Linguistics Vanguard}, page in press.

\bibitem[\protect\citename{Morrill}2000]{Morrill2000a}
Glyn Morrill.
\newblock 2000.
\newblock Incremental processing and acceptability.
\newblock {\em Computational Linguistics}, 25(3):319--338.

\bibitem[\protect\citename{Newson}2002]{Newson2002}
Roger Newson.
\newblock 2002.
\newblock Parameters behind ``nonparametric'' statistics: {Kendall}'s tau,
  {Somers' D} and median differences.
\newblock {\em Stata Journal}, 2(1):45--64.

\bibitem[\protect\citename{Pimentel \bgroup et al.\egroup }2020]{Pimentel2020a}
Tiago Pimentel, Brian Roark, and Ryan Cotterell.
\newblock 2020.
\newblock {Phonotactic complexity and its trade-offs}.
\newblock {\em Transactions of the Association for Computational Linguistics},
  8:1--18, 01.

\bibitem[\protect\citename{Pimentel \bgroup et al.\egroup }2021]{Pimentel2021a}
Tiago Pimentel, Irene Nikkarinen, Kyle Mahowald, Ryan Cotterell, and Damián
  Blasi.
\newblock 2021.
\newblock How (non-)optimal is the lexicon?
\newblock In {\em North American Chapter of the Association for Computational
  Linguistics}.

\bibitem[\protect\citename{Popescu \bgroup et al.\egroup }2013]{Popescu2013a}
Ioan-Iovitz Popescu, Sven Naumann, Emmerich Kelih, Andrij Rovenchak, Haruko
  Sanada, Anja Overbeck, Reginald Smith, Radek Cech, Panchanan Mohanty, Andrew
  Wilson, and Gabriel Altmann.
\newblock 2013.
\newblock Word length: Aspects and languages.
\newblock In {\em Studies in Quantitative Linguistics}, volume~3, pages
  224--281. RAM-Verlag, Lüdenscheid.

\bibitem[\protect\citename{Semple \bgroup et al.\egroup }2021]{Semple2021a}
Stuart Semple, Ramon {Ferrer-i-Cancho}, and Morgan Gustison.
\newblock 2021.
\newblock Linguistic laws in biology.
\newblock {\em Trends in Ecology and Evolution}, page in press.

\bibitem[\protect\citename{Shannon}1948]{Shannon1948}
Claude~E. Shannon.
\newblock 1948.
\newblock A mathematical theory of communication.
\newblock {\em Bell Systems Technical Journal}, 27:379--423\,623--656.

\bibitem[\protect\citename{Temperley and Gildea}2018]{Temperley2018a}
David Temperley and Daniel Gildea.
\newblock 2018.
\newblock Minimizing syntactic dependency lengths: {Typological/Cognitive}
  universal?
\newblock {\em Annual Review of Linguistics}, 4(1):67--80.

\bibitem[\protect\citename{Winter}2019]{Winter2019a}
Bodo Winter.
\newblock 2019.
\newblock {\em Statistics for linguists: an introdution using R}.
\newblock Routledge, New York \& London.

\bibitem[\protect\citename{Zeman \bgroup et al.\egroup }2017]{conll2017st_APS}
Daniel Zeman, Martin Popel, Milan Straka, Jan Haji{\v{c}}, Joakim Nivre, Filip
  Ginter, Juhani Luotolahti, Sampo Pyysalo, Slav Petrov, Martin Potthast,
  et~al.
\newblock 2017.
\newblock {C}o{NLL} 2017 shared task: Multilingual parsing from raw text to
  universal dependencies.
\newblock In {\em Proceedings of the {C}o{NLL} 2017 Shared Task: Multilingual
  Parsing from Raw Text to Universal Dependencies}, pages 1--19, Vancouver,
  Canada, August. Association for Computational Linguistics.

\bibitem[\protect\citename{Zeman \bgroup et al.\egroup }2020]{ud26_APS}
Daniel Zeman, Joakim Nivre, Mitchell Abrams, Elia Ackermann, No{\"e}mi Aepli,
  {\v Z}eljko Agi{\'c}, Lars Ahrenberg, Chika~Kennedy Ajede, Gabriel{\.e}
  Aleksandravi{\v c}i{\=u}t{\.e}, Lene Antonsen, et~al.
\newblock 2020.
\newblock Universal dependencies 2.6.
\newblock {LINDAT}/{CLARIAH}-{CZ} digital library at the Institute of Formal
  and Applied Linguistics ({{\'U}FAL}), Faculty of Mathematics and Physics,
  Charles University.

\bibitem[\protect\citename{Zipf}1949]{Zipf1949a}
George~Kingsley Zipf.
\newblock 1949.
\newblock {\em Human behaviour and the principle of least effort}.
\newblock Addison-Wesley, Cambridge (MA), USA.

\end{thebibliography}

\end{document}